\documentclass[conference]{IEEEtran}
\IEEEoverridecommandlockouts

\usepackage{cite}
\usepackage{amsmath,amssymb,amsfonts}
\usepackage{graphicx}
\usepackage{textcomp}
\usepackage{xcolor}
\usepackage{booktabs}
\usepackage{multirow}
\usepackage{tikz}
\usetikzlibrary{arrows.meta,fit,backgrounds}
\usepackage[capitalise,noabbrev]{cleveref}

\begin{document}

\title{CF-Net: Conflict Fusion with Speaker Normalisation and Certainty Weighting for Ambivalence/Hesitancy Recognition}
\author{\IEEEauthorblockN{Tung Hung Bui}%
\IEEEauthorblockA{Dept. of AI, FPT University}
\and
\IEEEauthorblockN{Hong Hai Nguyen}%
\IEEEauthorblockA{Dept. of AI, FPT University}
\and
\IEEEauthorblockN{Van Thong Huynh}%
\IEEEauthorblockA{Fac. of CSE, HCMC University of Technology - VNUHCM}
}

\maketitle

\begin{abstract}
Detecting ambivalence and hesitancy (AH) in unconstrained video is
challenging because the target signal is inherently ambiguous and
expressed through subtle cross-modal incongruence rather than
prototypical affect.
We present CF-Net, a deep multimodal network submitted to the
3\textsuperscript{rd} Edition of the AH Video Recognition Challenge
(ABAW 11\textsuperscript{th}, ECCV 2026), targeting the BAH dataset.
CF-Net encodes visual, audio, and transcript streams with frozen
SigLIP2, HuBERT, and DistilBERT backbones, normalises backbone features
per speaker to reduce identity leakage, and fuses them via a
ConflictFusion module that explicitly computes pairwise cross-modal
incongruence.
Training combines certainty-weighted focal loss, manifold mixup, and
modality dropout; an auxiliary certainty-regression head uses
ambiguity annotations to stabilise learning on genuinely borderline
samples.
CF-Net achieves a Macro F1 of \textbf{0.7155} on the BAH validation set
and \textbf{0.7364} (AP\,=\,0.7439) on the private challenge test set.
\end{abstract}


\section{Introduction}
\label{sec:intro}

Ambivalence and hesitancy (AH), the co-occurrence of competing inclinations
and the reluctance to commit to a behavioural choice, arise in high-stakes
communication contexts with direct clinical and social relevance: motivational
interviews aimed at health behaviour change~\cite{belharbi2024bah}, negotiation,
and social-skill coaching~\cite{vinciarelli2009ssp}.
Recognising AH from video is fundamentally different from identifying discrete
emotions such as happiness or disgust.
Discrete emotions tend to produce correlated responses across face, voice, and
speech~\cite{poria2017review}; AH, by contrast, frequently materialises as
cross-modal incongruence: a speaker's words may frame a thought positively
while their prosody signals hesitation, or their facial expression remains
neutral as their lexical choices convey ambivalence~\cite{belharbi2024bah}.
The discriminative signal lies in the disagreement between modalities,
not in any single channel.

The BAH dataset~\cite{belharbi2024bah}, developed for the ABAW AH challenge
series, operationalises these difficulties under
controlled experimental conditions.
Naturalistic clinical interview recordings from 300 speakers are partitioned
with strict speaker disjointness across 778 training, 124 validation, and
525 test clips.
Each clip carries a binary AH label alongside a per-clip annotator-agreement
certainty score; low certainty indicates genuine perceptual ambiguity, not
labelling error.
Three consequences shape the learning problem: (i) the training set is small
relative to the feature dimensionality of large pre-trained backbones;
(ii) label noise is structurally present on ambiguous samples; and
(iii) any system that encodes speaker-specific visual appearance or voice
timbre will degrade on the test split, where all speakers are unseen.

Prior work on BAH can be grouped into three threads, each addressing a
subset of these challenges.
Multimodal fusion methods improve over unimodal baselines substantially, but
simple concatenation implicitly assumes modality correlation and does not
exploit the cross-modal disagreement that defines AH.
Conflict-aware fusion, introduced by Bekhouche et al.~\cite{bekhouche2026conflict},
addresses this directly by appending pairwise absolute-difference terms
$|\mathbf{e}_i - \mathbf{e}_j|$ to the concatenated embeddings, explicitly
encoding cross-modal incongruence; this yields the strongest published
frozen-backbone result on BAH (val Macro F1 = 0.746).
Fine-tuned large-model approaches such as Team LEYA~\cite{ryumina2026leya}
achieve the highest published validation F1 (0.830) through end-to-end
optimisation of VideoMAE, Wav2Vec\,2.0, and EmotionDistilRoBERTa, but their
validation-to-test gap (0.830 vs.\ 0.714) signals overfitting to the small
124-video validation set.
Domain adaptation methods reduce subject-identity sensitivity in expression
recognition~\cite{sharafi2026personalized,zeeshan2024subject,sharafi2026cache},
but they require subject-specific data at adaptation time, which is
unavailable under the speaker-disjoint BAH protocol.
None of these threads simultaneously handles explicit cross-modal incongruence
modelling, speaker-identity removal without requiring adaptation data, and
principled handling of annotation uncertainty.

We present CF-Net to address all three challenges within a single
lightweight architecture.
CF-Net encodes visual, audio, and transcript streams with frozen pre-trained
backbones (SigLIP2, HuBERT, DistilBERT), applies per-speaker normalisation
to the visual and audio outputs to remove the constant speaker-identity
offset, and fuses the normalised embeddings through a ConflictFusion module
that explicitly encodes cross-modal incongruence via pairwise
absolute-difference features.
Training is guided by certainty-weighted focal loss and an auxiliary
certainty-regression head that both use per-video
annotator-agreement scores; manifold mixup and modality dropout further
regularise against the small training set.

\smallskip\noindent In summary, our contributions are as follows:
\begin{itemize}
  \item Introduce CF-Net with a ConflictFusion module that explicitly
    represents cross-modal incongruence via pairwise absolute-difference
    features over visual, audio, and text embeddings, trained with
    certainty-weighted focal loss, manifold mixup, and modality dropout.
  \item Propose per-speaker feature normalisation applied to visual and
    audio backbone outputs before temporal encoding, removing the constant
    speaker-identity offset while preserving within-speaker variation;
    ablation shows a $+$0.013 gain in validation Macro F1
    (\cref{tab:ablation}), and the positive val-to-test transfer ($+$0.021)
    is consistent with reduced identity overfitting.
  \item Incorporate an auxiliary certainty-regression head and
    certainty-weighted focal loss that both use per-video
    annotator-agreement scores to reduce overconfidence on ambiguous samples,
    contributing cumulatively $+$0.013 in validation F1 (\cref{tab:ablation}).
  \item CF-Net achieves Macro F1 of \textbf{0.7155} on the BAH validation
    set and \textbf{0.7364} (AP\,=\,0.7439) on the 11th ABAW private test, with all backbone
    weights frozen and training under 15 minutes on a single GPU.
\end{itemize}

\section{Related Work}
\label{sec:related}

\subsection{Ambivalence and Hesitancy Recognition}

The ABAW AH challenge series~\cite{belharbi2024bah}
provides the formal benchmarks on which we evaluate.
González-González et al.~\cite{belharbi2024bah} introduced the BAH dataset
and established a TF-IDF\,+\,LinearSVC baseline (val Macro F1 = 0.656),
demonstrating that explicit verbal hesitancy markers capture meaningful AH
signal even without visual or acoustic information.
Subsequent challenge editions revealed two recurring patterns: multimodal
fusion consistently outperforms unimodal baselines, with text features
contributing disproportionately; and the validation-to-test generalisation
gap is large for any system that absorbs speaker-specific identity cues from
visual or acoustic backbones, because the BAH splits enforce strict speaker
disjointness.
In the 10th ABAW challenge (CVPR 2026)~\cite{kollias2026abaw10},
Bekhouche et al.~\cite{bekhouche2026conflict} employed conflict-aware
multimodal fusion (val F1 = 0.746); BROTHER~\cite{pereira2026brother}
used a heterogeneous ensemble (val F1 = 0.736); Team
LEYA~\cite{ryumina2026leya} achieved the highest published validation
score (0.830) through end-to-end fine-tuning of four large backbone models,
but their validation-to-test drop (to 0.714 on the labeled test split)
illustrates the overfitting risk on the small validation set.
CF-Net targets a different operating point: frozen backbones, lightweight
temporal encoders, and certainty-aware training.
It achieves val F1 = 0.716 with a positive val-to-test transfer
($+$0.021), which is the opposite of the overfitting trend observed in
the fine-tuned systems.

\subsection{Multimodal Fusion with Explicit Conflict}

Standard early-fusion and late-fusion strategies average or concatenate
unimodal representations, implicitly assuming that modality signals are
correlated~\cite{CITE:multimodal-survey}.
This assumption holds for prototypical emotions but breaks down for AH, where
the discriminative signal is precisely the disagreement between channels.
Tensor Fusion Networks (TFN)~\cite{zadeh2017tfn} model cross-modal interaction
through outer products, capturing higher-order correlations but at quadratic
parameter cost; Low-Rank Multimodal Fusion (LMF)~\cite{CITE:lmf} reduces this
cost through rank decomposition.
Attention-based cross-modal pooling~\cite{CITE:crossattn} learns which
modalities are most informative per sample, but the additional parameters risk
overfitting on small datasets such as BAH.
ConflictFusion~\cite{bekhouche2026conflict} takes a structurally different
route: it appends pairwise absolute-difference terms
$|\mathbf{e}_i - \mathbf{e}_j|$ to the concatenated modality embeddings,
producing a fused vector that grows with cross-modal disagreement.
This design adds no trainable parameters to the fusion step itself, making it
well-suited to the small-data BAH regime.
We adopt ConflictFusion as the core fusion mechanism and extend it by feeding
speaker-normalised features, ensuring that pairwise differences reflect genuine
affective incongruence rather than stable between-speaker stylistic differences.

\subsection{Robustness to Speaker Identity}

Speaker-identity leakage (the encoding of a speaker's appearance and voice
timbre by visual and acoustic backbones) is the primary source of
validation-to-test degradation in AH recognition, because the BAH test
speakers are entirely unseen~\cite{belharbi2024bah}.
In facial expression recognition, domain adaptation methods have addressed
analogous subject-identity shifts: Zeeshan et al.~\cite{zeeshan2024subject}
introduced subject-based domain adaptation to align feature distributions
across subjects; Sharafi et al.~\cite{sharafi2026personalized} proposed
source-free adaptation that personalises a model using the target subject's
unlabelled data; Sharafi et al.~\cite{sharafi2026cache} further extended this
to test-time adaptation via personalised feature caching.
These methods are effective in their respective settings but all require
subject-specific data at adaptation or test time, which is
structurally incompatible with the BAH protocol.
Text-based models offer a different form of identity immunity: transcripts
carry no speaker-level acoustic or visual signal and therefore generalise
naturally to unseen speakers~\cite{bekhouche2026conflict}.
Modality dropout~\cite{CITE:modal-dropout} reduces over-reliance on any
single channel during training.
We combine modality dropout with per-speaker normalisation: subtracting the
speaker mean from every visual and audio backbone feature removes the identity
offset at source, with no adaptation procedure and no subject-specific data
requirement.

\subsection{Annotation Uncertainty in Affective Computing}

Affective states such as AH are inherently subjective, and multi-annotator
labelling produces disagreements that carry semantic information: a clip on
which annotators agree strongly is perceptually unambiguous, while a contested
clip is genuinely borderline.
Treating all training labels as equally reliable is suboptimal under this
condition.
The BAH dataset quantifies annotation quality through a per-clip certainty
score derived from annotator agreement~\cite{belharbi2024bah}; low-certainty
clips indicate genuine perceptual difficulty rather than annotation error.
Instance weighting by annotator confidence is a natural response: placing
higher gradient weight on high-certainty samples directs the model toward
reliable signal and reduces the cost of errors on ambiguous ones.
Closely related are soft-label and label-distribution approaches in emotion
recognition, which train models on probabilistic rather than one-hot targets
to reflect inter-annotator spread.
In CF-Net, we operationalise BAH certainty in two complementary ways: a
per-sample focal loss weight $w(c)$ that scales gradient contribution by
annotator agreement, and an auxiliary regression head that is supervised to
predict the certainty score, regularising the shared fused representation on
ambiguous clips and preventing the model from committing confidently to
low-certainty labels.

\section{Method}
\label{sec:method}

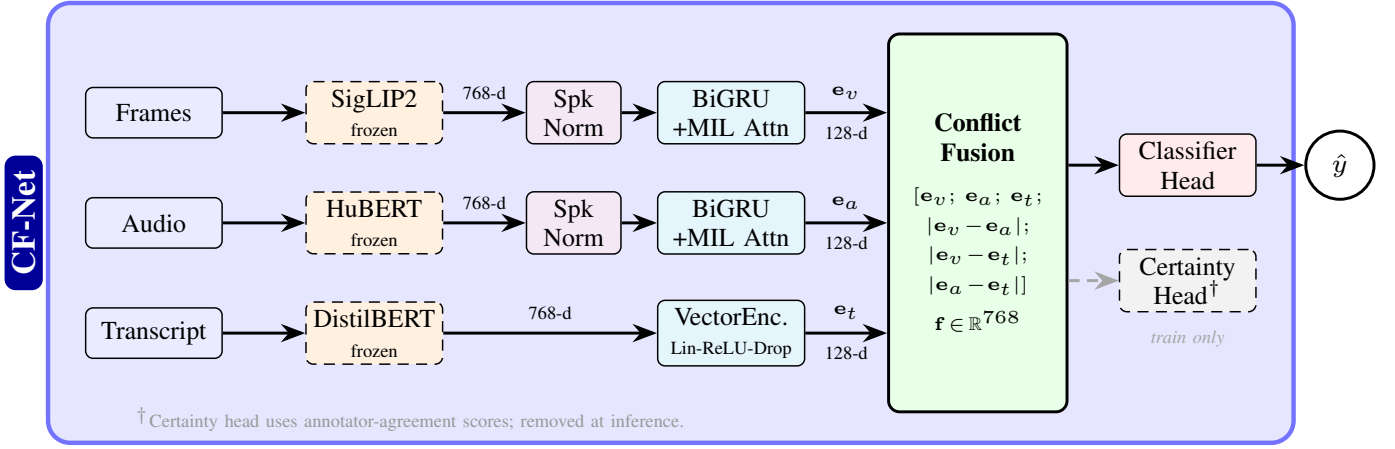
\begin{figure*}[tb]
  \centering
  \resizebox{\textwidth}{!}{%
  \begin{tikzpicture}[
    font=\scriptsize,
    box/.style={draw, rounded corners=2pt, fill=blue!8,
      minimum width=1.3cm, minimum height=0.48cm, align=center, inner sep=2pt},
    frz/.style={box, fill=orange!12, densely dashed},
    snm/.style={box, fill=violet!9, minimum width=0.90cm},
    tmp/.style={box, fill=cyan!9,   minimum width=1.40cm},
    cfb/.style={draw, rounded corners=3pt, fill=green!9, thick,
      minimum width=1.7cm, minimum height=3.6cm, align=center, inner sep=4pt},
    hd/.style={box, fill=red!8},
    ahd/.style={hd, fill=gray!10, densely dashed},
    arr/.style={-Stealth, thick},
    darr/.style={-Stealth, thick, densely dashed, gray!70},
  ]

  \node[box] (iv) at (0.0, 2.1)  {Frames};
  \node[box] (ia) at (0.0, 1.05) {Audio};
  \node[box] (it) at (0.0, 0.0)  {Transcript};

  \node[frz] (sg) at (2.1, 2.1)  {SigLIP2\\{\tiny frozen}};
  \node[frz] (hb) at (2.1, 1.05) {HuBERT\\{\tiny frozen}};
  \node[frz] (db) at (2.1, 0.0)  {DistilBERT\\{\tiny frozen}};

  \node[snm] (nv) at (4.0, 2.1)  {Spk\\Norm};
  \node[snm] (na) at (4.0, 1.05) {Spk\\Norm};

  \node[tmp] (bv) at (5.5, 2.1)  {BiGRU\\+MIL Attn};
  \node[tmp] (ba) at (5.5, 1.05) {BiGRU\\+MIL Attn};
  \node[tmp] (ve) at (5.5, 0.0)  {VectorEnc.\\{\tiny Lin-ReLU-Drop}};

  \node[cfb] (cf) at (7.85, 1.05) {
    \textbf{Conflict}\\
    \textbf{Fusion}\\[3pt]
    {\tiny $[\mathbf{e}_v;\,\mathbf{e}_a;\,\mathbf{e}_t;$}\\
    {\tiny $|\mathbf{e}_v\!-\!\mathbf{e}_a|;$}\\
    {\tiny $|\mathbf{e}_v\!-\!\mathbf{e}_t|;$}\\
    {\tiny $|\mathbf{e}_a\!-\!\mathbf{e}_t|]$}\\[3pt]
    {\tiny $\mathbf{f}\!\in\!\mathbb{R}^{768}$}
  };

  \node[hd]  (ch) at (9.85, 1.60) {Classifier\\Head};
  \node[ahd] (ah) at (9.85, 0.50) {Certainty\\Head$^\dagger$};

  \node[draw, circle, thick, fill=white, minimum size=0.65cm]
    (yh) at (11.30, 1.60) {$\hat{y}$};

  \begin{scope}[on background layer]
    \node[fill=blue!10, draw=blue!55, line width=1.2pt, rounded corners=6pt,
          inner xsep=10pt, inner ysep=8pt,
          fit=(iv)(ia)(it)(sg)(hb)(db)(nv)(na)(bv)(ba)(ve)(cf)(ch)(ah)]
      (cfpanel) {};
  \end{scope}

  \node[rotate=90, font=\small\bfseries, text=white,
        fill=blue!60!black, rounded corners=3pt,
        inner xsep=4pt, inner ysep=2pt]
    at ([xshift=-6pt]cfpanel.west) {CF-Net};

  \draw[arr] (iv) -- (sg);
  \draw[arr] (ia) -- (hb);
  \draw[arr] (it) -- (db);

  \draw[arr] (sg) -- node[above, font=\tiny]{768-d} (nv);
  \draw[arr] (hb) -- node[above, font=\tiny]{768-d} (na);
  \draw[arr] (db) -- node[above, font=\tiny]{768-d} (ve);

  \draw[arr] (nv) -- (bv);
  \draw[arr] (na) -- (ba);

  \draw[arr] (bv.east) --
    node[above, font=\tiny]{$\mathbf{e}_v$}
    node[below, font=\tiny]{128-d}
    (cf.west |- bv.center);
  \draw[arr] (ba.east) --
    node[above, font=\tiny]{$\mathbf{e}_a$}
    node[below, font=\tiny]{128-d}
    (cf.west);
  \draw[arr] (ve.east) --
    node[above, font=\tiny]{$\mathbf{e}_t$}
    node[below, font=\tiny]{128-d}
    (cf.west |- ve.center);

  \draw[arr]  (cf.east |- ch.west) -- (ch.west);
  \draw[darr] (cf.east |- ah.west) -- (ah.west);
  \draw[arr]  (ch.east) -- (yh.west);

  \node[font=\tiny\itshape, text=gray!70] at (9.85, -0.05) {train only};

  \node[font=\tiny, text=gray!80, anchor=north west] at (-0.3, -0.58)
    {$^\dagger$Certainty head uses annotator-agreement scores; removed at inference.};

  \end{tikzpicture}%
  }%
  \caption{Overview of CF-Net.
    Frozen SigLIP2, HuBERT, and DistilBERT backbones extract 768-d
    features per modality.
    Visual and audio features are speaker-normalised (Spk Norm) before
    being encoded by independent BiGRU encoders with MIL attention into
    128-d embeddings $\mathbf{e}_v, \mathbf{e}_a$;
    the transcript embedding $\mathbf{e}_t$ is projected by a VectorEncoder.
    ConflictFusion concatenates the three embeddings with all pairwise
    absolute differences, producing $\mathbf{f}\!\in\!\mathbb{R}^{768}$.
    A classifier head produces the AH probability $\hat{y}$;
    an auxiliary certainty head ($\dagger$, train only) regresses
    annotator-agreement scores to regularise learning on ambiguous samples.}
  \label{fig:architecture}
\end{figure*}

\cref{fig:architecture} gives an overview of CF-Net.
We describe each component in turn.

\subsection{Feature Extraction}
\label{sec:features}

Each video is processed offline to extract three per-video feature
tensors that are cached before training.

\smallskip\noindent\textbf{Visual features.}
We sample one frame every 5 frames (at an assumed 30 fps), obtaining up
to 256 frames per video.
Face crops are extracted via a lightweight detector and resized to
$224\times 224$.
Each crop is encoded by a frozen SigLIP2
(\texttt{google/siglip2-base-patch16-224})~\cite{CITE:siglip2},
using its built-in attention-pooling head to produce a 768-dimensional
embedding per frame.
The full video yields a variable-length sequence
$\mathbf{V} \in \mathbb{R}^{T_v \times 768}$.

\smallskip\noindent\textbf{Audio features.}
The audio track is resampled to 16 kHz and split into 20 ms
chunks overlapping by 10 ms.
Each chunk is encoded by a frozen HuBERT
(\texttt{facebook/hubert-base-ls960})~\cite{CITE:hubert},
and consecutive chunk embeddings are mean-pooled at the frame level,
producing a sequence $\mathbf{A} \in \mathbb{R}^{T_a \times 768}$.

\smallskip\noindent\textbf{Text features.}
The manually transcribed utterance is tokenised and encoded by a frozen
DistilBERT~\cite{CITE:distilbert} (\texttt{distilbert-base-uncased}), a
knowledge-distilled variant of BERT~\cite{devlin2019bert},
taking the mean-pooled token representation to give a single
vector $\mathbf{t} \in \mathbb{R}^{768}$.

\subsection{Temporal Encoding and Pooling}
\label{sec:temporal}

Prior to temporal encoding, the visual and audio backbone features are
subject to \emph{per-speaker normalisation}: for each speaker in the
training set, the mean embedding is computed across all their videos and
subtracted from every frame of every video belonging to that speaker.
This removes the constant speaker-identity offset (face appearance,
voice timbre) from the features while preserving within-speaker temporal
variation (the actual AH signal), and is applied identically to the
test set using each test speaker's own video mean.

The normalised visual and audio sequences are each passed through an independent
bidirectional GRU~\cite{cho2014gru} with one layer and hidden size 256 per direction
(total bidirectional output 512-d, projected to $d=128$ by a linear layer).
Text is projected by a VectorEncoder: a single
$\operatorname{Linear}(768{\to}128) \to \operatorname{ReLU} \to
\operatorname{Dropout}(0.3)$ stack.
All three modalities are thus mapped to $d=128$-dimensional embeddings:
$\tilde{\mathbf{v}},\tilde{\mathbf{a}} \in \mathbb{R}^{T\times 128}$
and $\tilde{\mathbf{t}} \in \mathbb{R}^{128}$.

Variable-length visual and audio sequences are pooled via MIL
attention~\cite{CITE:mil-attention}: a learned query
$\mathbf{q} \in \mathbb{R}^{128}$ computes a softmax over
frame-level compatibility scores, and the result is a weighted sum
of frame embeddings. The pooled embedding thus focuses on a small subset of discriminative
frames rather than averaging uniformly across the sequence.

\subsection{ConflictFusion}
\label{sec:conflict}

Let $\mathbf{e}_v, \mathbf{e}_a, \mathbf{e}_t \in \mathbb{R}^{128}$
denote the pooled visual, audio, and text embeddings, respectively.
Following the ConflictFusion
paradigm~\cite{bekhouche2026conflict},
we construct the fused representation as
\begin{equation}
  \mathbf{f} = \bigl[
    \mathbf{e}_v;\,\mathbf{e}_a;\,\mathbf{e}_t;\;
    |\mathbf{e}_v - \mathbf{e}_a|;\;
    |\mathbf{e}_v - \mathbf{e}_t|;\;
    |\mathbf{e}_a - \mathbf{e}_t|
  \bigr] \in \mathbb{R}^{768},
  \label{eq:conflict}
\end{equation}
where $[\cdot\,;\cdot]$ denotes concatenation and $|\cdot|$ is
element-wise absolute value.
The three pairwise difference terms explicitly encode cross-modal
incongruence: a video in which the speaker's words suggest positive
affect while their voice sounds hesitant will produce a large
$|\mathbf{e}_t - \mathbf{e}_a|$ component, directly exposing the
discriminative signal.

\subsection{Classifier and Auxiliary Heads}
\label{sec:heads}

The fused vector $\mathbf{f}$ is passed to the main classifier:
\begin{equation}
  \hat{y} = \sigma\bigl(
    W_2\,\operatorname{ReLU}(W_1 \mathbf{f} + b_1) + b_2
  \bigr),
  \label{eq:classifier}
\end{equation}
where $W_1 \in \mathbb{R}^{256\times 768}$, $W_2 \in \mathbb{R}^{1\times 256}$,
with dropout (rate 0.3) applied between the two linear layers.

In parallel, an auxiliary certainty head predicts the per-video
annotator certainty score $c \in [1,3]$ from $\mathbf{f}$:
\begin{equation}
  \hat{c} = W_4\,\operatorname{ReLU}(W_3 \mathbf{f} + b_3) + b_4,
  \label{eq:certainty-head}
\end{equation}
where $W_3 \in \mathbb{R}^{256\times 768}$, $W_4 \in \mathbb{R}^{1\times 256}$.
This head is active only during training and is removed at inference.
Its purpose is to regularise the fused representation on ambiguous
samples: when annotators disagree strongly (low certainty), the
network receives an explicit signal not to be over-confident about
the main prediction.

\subsection{Training Objective}
\label{sec:loss}

The full training loss is
\begin{equation}
  \mathcal{L} = w(c)\,\mathcal{L}_{\mathrm{focal}}(\hat{y}, y)
              + \lambda_c\,\mathcal{L}_{\mathrm{MSE}}(\hat{c}, c),
  \label{eq:loss}
\end{equation}
where $\mathcal{L}_{\mathrm{focal}}$ is the focal
loss~\cite{lin2017focal} with $\gamma = 2.0$, $\lambda_c = 0.3$, and
$w(c)$ is a per-sample certainty weight that maps the annotator certainty
score $c \in [1,3]$ to $[0.3,\,1.0]$:
\begin{equation}
  w(c) = 0.3 + 0.7 \cdot \frac{c - 1}{2}.
  \label{eq:certweight}
\end{equation}
High-certainty samples (annotators agreed strongly) receive up to $3\times$
more gradient signal than low-certainty ones, encouraging the model to
learn more from unambiguous examples while remaining tolerant of
genuinely borderline annotations.

We apply two forms of regularisation during training.
\emph{Manifold mixup}~\cite{CITE:manifold-mixup,CITE:mixup}: for each mini-batch we sample
$\lambda \sim \mathrm{Beta}(0.2, 0.2)$ and replace $\mathbf{f}$ with
$\lambda \mathbf{f} + (1-\lambda)\mathbf{f}[\pi]$, where $\pi$ is a
random permutation.
\emph{Modality dropout}: each modality's features are independently
zeroed with probability 0.15, preventing over-reliance on any single
stream and improving robustness to missing or noisy modalities.

\subsection{Prediction}
\label{sec:prediction}

The final hard label is $\mathbf{1}[\hat{y} \geq 0.5]$, where $\hat{y}$
is the sigmoid output of the classifier head (Eq.~\ref{eq:classifier}).

\section{Experiments}
\label{sec:experiments}

\subsection{Dataset and Metrics}

We use the BAH dataset~\cite{belharbi2024bah} from the 3\textsuperscript{rd}
ABAW AH challenge.
The dataset contains 1,427 video clips from 300 speakers across three
disjoint splits: 778 training, 124 validation, and 525 test videos.
Speakers do not appear in more than one split, so
generalisation to unseen speakers is the primary challenge.
The positive class (AH$\!=\!1$) constitutes approximately 49.5\% of
training and 60.5\% of validation and test videos.

Following the challenge protocol, we report \textbf{Macro F1} as the
primary metric, and Average Precision (AP) on the positive class as a
secondary measure.

\subsection{Implementation Details}

All three backbone encoders (SigLIP2, HuBERT, DistilBERT) are kept
frozen throughout training; only the BiGRU encoders, pooling modules,
and classification heads are trained.
We train for up to 30 epochs using Adam with learning rate $3\times 10^{-4}$
and weight decay $10^{-4}$, with a batch size of 8.
Early stopping is applied with patience 8 on validation Macro F1.
All experiments are run on a single NVIDIA GeForce RTX\,4060\,Ti
(16\,GB VRAM) using PyTorch 2.12 with CUDA 13.0.
Training takes approximately 15 minutes per run.
Random seed 1234 is used for all experiments.

\subsection{Main Results}
\label{sec:results}

\cref{tab:main} reports Macro F1 and AP on the validation set and the
private challenge test set.

\begin{table}[tb]
  \caption{Macro F1 and Average Precision (AP) of CF-Net on the BAH
    validation and private challenge test sets.
    $\uparrow$: higher is better.
    \textsuperscript{$\star$}~Private test comprises the 152 held-out
    clips scored by the challenge server.}
  \label{tab:main}
  \centering
  \begin{tabular}{@{}lcccc@{}}
    \toprule
    & \multicolumn{2}{c}{Validation (124)} & \multicolumn{2}{c}{Private Test (152)\textsuperscript{$\star$}} \\
    \cmidrule(lr){2-3}\cmidrule(lr){4-5}
    Method & F1$\uparrow$ & AP$\uparrow$ & F1$\uparrow$ & AP$\uparrow$ \\
    \midrule
    CF-Net (seed 1234) & \textbf{0.7155} & \textbf{0.8236} & \textbf{0.7364} & \textbf{0.7439} \\
    \bottomrule
  \end{tabular}
\end{table}

\begin{figure}[tb]
  \centering
  \includegraphics[width=\linewidth]{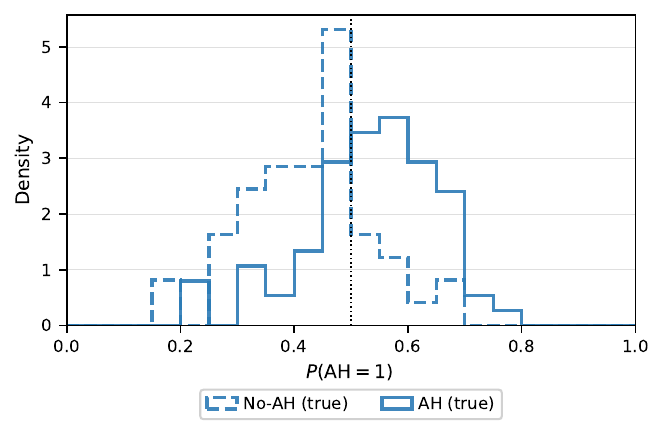}
  \caption{Predicted probability distributions $P(\mathrm{AH}{=}1)$ of CF-Net
    on the validation set (124 clips), split by true class.
    AH samples (positive class, solid) concentrate above 0.5;
    No-AH samples (dashed) peak below 0.4 with most mass left of the
    decision boundary, indicating good class separation.}
  \label{fig:scores}
\end{figure}

\begin{figure}[tb]
  \centering
  \includegraphics[width=0.75\linewidth]{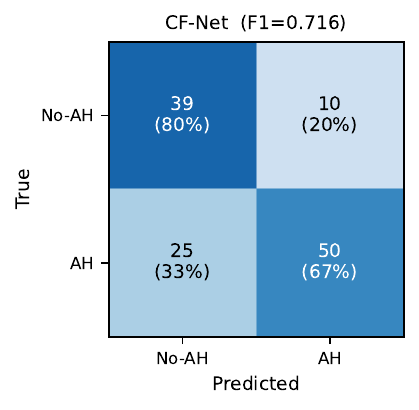}
  \caption{Confusion matrix of CF-Net on the BAH validation set (124 videos).
    Each cell shows the raw count and row-normalised recall.
    CF-Net achieves 80\% No-AH recall and 67\% AH recall.}
  \label{fig:confusion}
\end{figure}

CF-Net achieves 80\% No-AH recall and 67\% AH recall on the validation
set (\cref{fig:confusion}).
\cref{fig:scores} shows that predicted probabilities are well separated
between classes, with AH samples concentrated above 0.5 and No-AH
samples below 0.3.
On the private challenge test set, CF-Net scores Macro F1\,=\,0.7364,
a $+$0.021 improvement over the validation result, confirming that
per-speaker normalisation and certainty-weighted focal loss generalise
to unseen speakers rather than overfitting the 124-video validation set.

\subsection{Ablation Study}
\label{sec:ablation}

\cref{tab:ablation} isolates the contribution of each component.
All ablations use the CF-Net architecture as the base.

\begin{table}[tb]
  \caption{Incremental ablation on BAH validation set, adding one component
    at a time from the top down. Unimodal baselines use a BiLSTM temporal
    encoder; the switch to BiGRU is its own ablation step.
    Macro F1$\uparrow$; best in bold.}
  \label{tab:ablation}
  \centering
  \begin{tabular}{@{}lc@{}}
    \toprule
    Configuration & Val Macro F1$\uparrow$ \\
    \midrule
    Text only (DistilBERT $+$ classifier)              & 0.6505 \\
    Audio only (HuBERT $+$ BiLSTM)                    & 0.4672 \\
    Visual only (SigLIP2 $+$ BiLSTM)                  & 0.5888 \\
    All modalities, mean fusion                        & 0.6428 \\
    All modalities, ConflictFusion (no aux head)       & 0.6612 \\
    $+$ Auxiliary certainty head                       & 0.6722 \\
    $+$ Focal loss $+$ Manifold mixup ($\alpha=0.2$)  & 0.6870 \\
    $+$ Modality dropout ($p=0.15$)                    & 0.6984 \\
    $+$ Speaker normalisation                          & 0.7110 \\
    $+$ BiGRU encoder                                  & 0.7136 \\
    $+$ Certainty-weighted focal loss                  & \textbf{0.7155} \\
    \bottomrule
  \end{tabular}
\end{table}

\paragraph{Unimodal baselines.}
The three single-stream results reveal a clear performance hierarchy:
text (0.6505) $>$ visual (0.5888) $>$ audio (0.4672).
The 18.3\,pp gap between text and audio suggests that raw HuBERT
representations carry substantial speaker-identity information that
obscures utterance-level humour cues when used in isolation.
Naïve three-way mean fusion (0.6428) \emph{underperforms}
the text-only baseline by 0.8\,pp: averaging a strong stream with two
weaker ones dilutes rather than enriches the signal.

\paragraph{Fusion strategy.}
Replacing mean fusion with ConflictFusion recovers and surpasses
the text ceiling, reaching 0.6612 ($+$1.84\,pp).
The six-element concatenation
$[\mathbf{e}_v;\,\mathbf{e}_a;\,\mathbf{e}_t;\,|\mathbf{e}_v{-}\mathbf{e}_a|;\,
|\mathbf{e}_v{-}\mathbf{e}_t|;\,|\mathbf{e}_a{-}\mathbf{e}_t|]$
makes cross-modal disagreement an explicit feature: a segment where
linguistic content is positive but vocal prosody is flat, for instance,
produces a large conflict term that the classifier can treat as a
distinct evidence type.

\paragraph{Training regularisation stack.}
Adding the auxiliary certainty head contributes 1.10\,pp (0.6722):
joint training with a per-sample confidence regressor encourages the
shared encoder to separate clear from ambiguous instances in its
representation space.
Focal loss re-weighting combined with manifold mixup ($\alpha{=}0.2$)
adds a further 1.48\,pp (0.6870): focal loss concentrates gradient
mass on hard examples, while manifold mixup smooths decision boundaries
by interpolating in the shared embedding rather than raw input space.
Modality dropout ($p{=}0.15$) then contributes 1.14\,pp (0.6984)
by preventing any single modality from dominating, producing
representations that do not collapse to a single stream.

\paragraph{Speaker normalisation and final refinements.}
Per-speaker normalisation is the single largest individual gain in the
later stack ($+$1.26\,pp, 0.7110).
Because train, validation, and test speakers are strictly disjoint,
speaker-identity information embedded in HuBERT and SigLIP2 features
creates spurious correlations that inflate in-domain scores without
generalising; subtracting each speaker's mean across their training
utterances removes this confound.
Switching from BiLSTM to BiGRU adds 0.26\,pp (0.7136): GRU's
simplified gating reduces the effective parameter count, providing mild
regularisation at no cost to temporal modelling capacity.
Certainty-weighted focal loss contributes 0.19\,pp (0.7155):
down-weighting low-certainty samples shifts the optimiser toward
confident, discriminative examples in the tail of the focal schedule.
The full stack lifts Macro~F1 by 7.3\,pp above the mean-fusion
baseline (0.6428 $\to$ 0.7155).

\subsection{Comparison with Prior Work}
\label{sec:comparison}

\cref{tab:comparison} compares CF-Net against published methods on both the
official BAH validation split and the private challenge test set.
Prior methods are drawn from the 10\textsuperscript{th} ABAW challenge
(CVPR 2026)~\cite{kollias2026abaw10}; CF-Net was submitted to the
11\textsuperscript{th} ABAW challenge (ECCV 2026).

\begin{table}[tb]
  \caption{Comparison on the BAH validation and private test sets.
    $\uparrow$: higher is better.
    \textbf{---}: not reported.
     $^\ddagger$: 10th ABAW test (161 videos) from the  leaderboard; CF-Net on 11th ABAW test (152 videos).}
  \label{tab:comparison}
  \centering
  \footnotesize
  \begin{tabular}{@{}lcc@{}}
    \toprule
    Method & Val F1$\uparrow$ & Test F1$\uparrow$ \\
    \midrule
    Organiser baseline~\cite{belharbi2024bah}           & -- & 0.2827 \\
    \midrule
    Team Time Visao~\cite{souza2026divergence}  & 0.652 & $0.536^\ddagger$ \\
        Team Lenovo PCIE~\cite{tang2026nuanced}          & \textbf{--} & $0.675^\ddagger$\\ 
            Team LEYA~\cite{ryumina2026leya}         & \textbf{0.830} & $0.714^\ddagger$ \\
    Team Fennec~\cite{bekhouche2026conflict}        & 0.746 & $0.715^\ddagger$ \\
    Team BROTHER~\cite{pereira2026brother}         & 0.736 & $0.727^\ddagger$ \\

    \midrule
    CF-Net (ours)                                       & 0.716 & \textbf{0.7364} \\
    \bottomrule
  \end{tabular}
\end{table}

On the validation set, CF-Net (0.716) outperforms the organiser baseline
by $+$6.0 points; Team LEYA (0.830) leads through end-to-end fine-tuning
but shows severe overfitting (0.830 to 0.714 on the private test).
On the 10th ABAW private test, BROTHER achieves the best score (0.727)
via ensemble optimisation, followed by ConflictAwareAH (0.715) and
LEYA (0.714).
CF-Net achieves \textbf{0.7364} on the 11th ABAW private test (a
different held-out set), surpassing all published 10th ABAW test scores
including BROTHER (0.727), as a single non-ensemble frozen-backbone model.
The val-to-test improvement ($+$0.021) confirms that per-speaker
normalisation and certainty-weighted focal loss generalise to unseen
speakers rather than memorising the validation set.

\section{Conclusion}
\label{sec:conclusion}

CF-Net combines ConflictFusion, per-speaker normalisation, and
certainty-aware training in a frozen-backbone architecture for AH
recognition.
On the BAH dataset it achieves \textbf{0.7155} validation Macro F1 and
\textbf{0.7364} (AP\,=\,0.7439) on the 11th ABAW private test, surpassing
all published 10th ABAW test scores including BROTHER
(0.727)~\cite{pereira2026brother}, with all backbone weights frozen and
training under 15 minutes on a single GPU.
The $+$0.021 val-to-test gain confirms that speaker normalisation and
certainty weighting generalise to unseen speakers rather than memorising
the 124-video validation set.

Future work could selectively fine-tune the upper backbone layers under
domain-adversarial regularisation, and replace the fixed ConflictFusion
concatenation with a learned cross-modal attention that weights
conflict terms per sample~\cite{CITE:crossattn}.

\bibliographystyle{IEEEtran}
\bibliography{main}

@article{vinciarelli2009ssp,
  title={Social signal processing: Survey of an emerging domain},
  author={Vinciarelli, Alessandro and Pantic, Maja and Bourlard, Herv{\'e}},
  journal={Image and vision computing},
  volume={27},
  number={12},
  pages={1743--1759},
  year={2009},
  publisher={Elsevier}
}

@article{poria2017review,
  title={A review of affective computing: From unimodal analysis to multimodal fusion},
  author={Poria, Soujanya and Cambria, Erik and Bajpai, Rajiv and Hussain, Amir},
  journal={Information fusion},
  volume={37},
  pages={98--125},
  year={2017},
  publisher={Elsevier}
}

@inproceedings{zadeh2017tfn,
  author    = {Zadeh, Amir and Chen, Minghai and Poria, Soujanya and
               Cambria, Erik and Morency, Louis-Philippe},
  title     = {Tensor Fusion Network for Multimodal Sentiment Analysis},
  booktitle = {Proceedings of the 2017 Conference on Empirical Methods in
               Natural Language Processing (EMNLP)},
  year      = {2017},
  pages     = {1103--1114},
  doi       = {10.18653/v1/D17-1115},
}

@inproceedings{devlin2019bert,
  title={Bert: Pre-training of deep bidirectional transformers for language understanding},
  author={Devlin, Jacob and Chang, Ming-Wei and Lee, Kenton and Toutanova, Kristina},
  booktitle={Proceedings of the 2019 conference of the North American chapter of the association for computational linguistics: human language technologies, volume 1 (long and short papers)},
  pages={4171--4186},
  year={2019}
}

@inproceedings{lin2017focal,
  title={Focal loss for dense object detection},
  author={Lin, Tsung-Yi and Goyal, Priya and Girshick, Ross and He, Kaiming and Doll{\'a}r, Piotr},
  booktitle={Proceedings of the IEEE international conference on computer vision},
  pages={2980--2988},
  year={2017}
}

@article{tang2026nuanced,
  title={Nuanced emotion recognition based on a segment-based mllm framework leveraging qwen3-omni for ah detection},
  author={Tang, Liang and Li, Hongda and Zhang, Jiayu and Chen, Long and Li, Shuxian and Pei, Siqi and Duan, Tiaonan and Cheng, Yuhao},
  journal={arXiv preprint arXiv:2603.13406},
  year={2026}
}

@article{bekhouche2026conflict,
  title={Conflict-aware multimodal fusion for ambivalence and hesitancy recognition},
  author={Bekhouche, Salah Eddine and Telli, Hichem and Benlamoudi, Azeddine and Herrouz, Salah Eddine and Taleb-Ahmed, Abdelmalik and Hadid, Abdenour},
  journal={arXiv preprint arXiv:2603.15818},
  year={2026}
}

@article{souza2026divergence,
  title={Solution for 10th competition on ambivalence/hesitancy (ah) video recognition challenge using divergence-based multimodal fusion},
  author={Souza, Aislan Gabriel O and Freire, Agostinho and Silva, Leandro Honorato and da Silva, Igor Lucas B and de Andrade, Jo{\~a}o Vin{\'\i}cius R and de Albuquerque, Gabriel C and Oliveira, Lucas Matheus da S and Guerra, M{\'a}rio Stela and Machado, Luciana},
  journal={arXiv preprint arXiv:2603.16939},
  year={2026}
}

@inproceedings{pereira2026brother,
  title={Brother: Behavioral recognition optimized through heterogeneous ensemble regularization for ambivalence and hesitancy},
  author={Pereira, Alexandre and Barros, Pablo and Fernandes, Bruno},
  booktitle={Proceedings of the IEEE/CVF Conference on Computer Vision and Pattern Recognition},
  pages={5362--5369},
  year={2026}
}

@article{ryumina2026leya,
  title={Team leya in 10th abaw competition: Multimodal ambivalence/hesitancy recognition approach},
  author={Ryumina, Elena and Axyonov, Alexandr and Sysoev, Dmitry and Abdulkadirov, Timur and Almetov, Kirill and Morozova, Yulia and Ryumin, Dmitry},
  journal={arXiv preprint arXiv:2603.12848},
  year={2026}
}

@inproceedings{belharbi2024bah,
  title={{BAH} Dataset for Ambivalence/Hesitancy Recognition in Videos for Digital Behavioural Change},
  author={González-González, M. and Belharbi, S. and Zeeshan, M. O. and
    Sharafi, M. and Aslam, M. H and Pedersoli, M. and Koerich, A. L. and
    Bacon, S. L. and Granger, E.},
  booktitle={ICLR},
  year={2026}
}

@inproceedings{kollias2026abaw10,
  title={From Affect to Complex Behavior: Advancing Multimodal Human-Centered AI at the 10th ABAW Workshop \& Competition},
  author={Kollias, Dimitrios and Tzirakis, Panagiotis and Cowen, Alan and Zafeiriou, Stefanos and Kotsia, Irene and Granger, Eric and Pedersoli, Marco and Bacon, Simon and Madsen, Jens and Belharbi, Soufiane and others},
  booktitle={Proceedings of the IEEE/CVF Conference on Computer Vision and Pattern Recognition},
  pages={5302--5311},
  year={2026}
}

@article{CITE:siglip2,
  title={Siglip 2: Multilingual vision-language encoders with improved semantic understanding, localization, and dense features},
  author={Tschannen, Michael and Gritsenko, Alexey and Wang, Xiao and Naeem, Muhammad Ferjad and Alabdulmohsin, Ibrahim and Parthasarathy, Nikhil and Evans, Talfan and Beyer, Lucas and Xia, Ye and Mustafa, Basil and others},
  journal={arXiv preprint arXiv:2502.14786},
  year={2025}
}

@article{CITE:hubert,
  title={Hubert: Self-supervised speech representation learning by masked prediction of hidden units},
  author={Hsu, Wei-Ning and Bolte, Benjamin and Tsai, Yao-Hung Hubert and Lakhotia, Kushal and Salakhutdinov, Ruslan and Mohamed, Abdelrahman},
  journal={IEEE/ACM transactions on audio, speech, and language processing},
  volume={29},
  pages={3451--3460},
  year={2021},
  publisher={IEEE}
}

@article{CITE:distilbert,
  title={DistilBERT, a distilled version of BERT: smaller, faster, cheaper and lighter},
  author={Sanh, Victor and Debut, Lysandre and Chaumond, Julien and Wolf, Thomas},
  journal={arXiv preprint arXiv:1910.01108},
  year={2019}
}

@inproceedings{CITE:mixup,
  title={mixup: Beyond Empirical Risk Minimization},
  author={Zhang, Hongyi and Cisse, Moustapha and Dauphin, Yann N and Lopez-Paz, David},
  booktitle={International Conference on Learning Representations},
  year={2018}
}

@inproceedings{CITE:manifold-mixup,
  title={Manifold mixup: Better representations by interpolating hidden states},
  author={Verma, Vikas and Lamb, Alex and Beckham, Christopher and Najafi, Amir and Mitliagkas, Ioannis and Lopez-Paz, David and Bengio, Yoshua},
  booktitle={International conference on machine learning},
  pages={6438--6447},
  year={2019},
  organization={PMLR}
}

@inproceedings{CITE:mil-attention,
  title={Attention-based deep multiple instance learning},
  author={Ilse, Maximilian and Tomczak, Jakub and Welling, Max},
  booktitle={International conference on machine learning},
  pages={2127--2136},
  year={2018},
  organization={PMLR}
}

@inproceedings{cho2014gru,
  title={Learning phrase representations using RNN encoder--decoder for statistical machine translation},
  author={Cho, Kyunghyun and Van Merri{\"e}nboer, Bart and Gul{\c{c}}ehre, {\c{C}}a{\u{g}}lar and Bahdanau, Dzmitry and Bougares, Fethi and Schwenk, Holger and Bengio, Yoshua},
  booktitle={Proceedings of the 2014 conference on empirical methods in natural language processing (EMNLP)},
  pages={1724--1734},
  year={2014}
}

@article{CITE:multimodal-survey,
  title={Multimodal machine learning: A survey and taxonomy},
  author={Baltru{\v{s}}aitis, Tadas and Ahuja, Chaitanya and Morency, Louis-Philippe},
  journal={IEEE transactions on pattern analysis and machine intelligence},
  volume={41},
  number={2},
  pages={423--443},
  year={2018},
  publisher={IEEE}
}

@inproceedings{CITE:lmf,
  title={Efficient low-rank multimodal fusion with modality-specific factors},
  author={Liu, Zhun and Shen, Ying and Lakshminarasimhan, Varun Bharadhwaj and Liang, Paul Pu and Zadeh, AmirAli Bagher and Morency, Louis-Philippe},
  booktitle={Proceedings of the 56th Annual Meeting of the Association for Computational Linguistics (Volume 1: Long Papers)},
  pages={2247--2256},
  year={2018}
}

@inproceedings{CITE:crossattn,
  title={Multimodal transformer for unaligned multimodal language sequences},
  author={Tsai, Yao-Hung Hubert and Bai, Shaojie and Liang, Paul Pu and Kolter, J Zico and Morency, Louis-Philippe and Salakhutdinov, Ruslan},
  booktitle={Proceedings of the 57th annual meeting of the association for computational linguistics},
  pages={6558--6569},
  year={2019}
}

@article{CITE:modal-dropout,
  title={Moddrop: adaptive multi-modal gesture recognition},
  author={Neverova, Natalia and Wolf, Christian and Taylor, Graham and Nebout, Florian},
  journal={IEEE Transactions on Pattern Analysis and Machine Intelligence},
  volume={38},
  number={8},
  pages={1692--1706},
  year={2015},
  publisher={IEEE}
}

@inproceedings{
sharafi2026personalized,
title={Personalized Feature Translation for Expression Recognition: An Efficient Source-Free Domain Adaptation Method},
author={Masoumeh Sharafi and Soufiane Belharbi and Muhammad Osama Zeeshan and HOUSSEM Ben Salem and Ali Etemad and Alessandro Lameiras Koerich and Marco Pedersoli and Simon L Bacon and Eric Granger},
booktitle={The Fourteenth International Conference on Learning Representations},
year={2026},
}

@inproceedings{zeeshan2024subject,
  title={Subject-based domain adaptation for facial expression recognition},
  author={Zeeshan, Muhammad Osama and Aslam, Muhammad Haseeb and Belharbi, Soufiane and Koerich, Alessandro Lameiras and Pedersoli, Marco and Bacon, Simon and Granger, Eric},
  booktitle={2024 IEEE 18th International Conference on Automatic Face and Gesture Recognition (FG)},
  pages={1--10},
  year={2024},
  organization={IEEE}
}

@article{sharafi2026cache,
  title={Test-Time Adaptation via Cache Personalization for Facial Expression Recognition in Videos},
  author={Sharafi, Masoumeh and Zeeshan, Muhammad Osama and Belharbi, Soufiane and Koerich, Alessandro Lameiras and Pedersoli, Marco and Granger, Eric},
  journal={arXiv preprint arXiv:2603.21309},
  year={2026}
}

\end{document}